В.В. Кромер

**Глоттохронологическая ретрогностика языковой системы**

В данной статье ставится задача ретрогностики (реконструкции предшествующего состояния) лингвистической системы, состоящей из определенного количества современных языков. Исходная информация содержится в матрице коэффициентов совпадений основного списка для языков системы. В результате реконструкции выявляется предшествующее состояние системы, рассматриваемой как набор взаимосвязанных изолектных цепей, где каждый существовавший в прошлом изолект представлен точкой одной из цепей. Отдельные изолекты связаны линиями дивергентного развития с изолектами других цепей либо с современными языками. Изолекты, связанные линиями дивергентного развития, считаются связанными отношениями «изолект-предок – изолект-потомок»; изолекты, связанные линиями изолектных цепей – синхронно существующими диалектами одного языка. Основные принципы реконструкции предшествующего состояния языковой системы восходят к идеям основателя глоттохронологии М. Сводеша [1]. Нами предложена модификация метода [2]. Ширину изолектных цепей и продолжительность дивергентного развития целесообразно определять в относительных единицах (единицах измерения лингвистического времени). В качестве лингвистической единицы измерения времени предлагается ввести сводеш в честь М. Сводеша и определить эту единицу как предел отношения $100 \lim_{c \to 1} \frac{\ln c}{c-1}$, где $c$ – коэффициент совпадения основных списков. Считая значение $c = 0{,}99$ достаточно близким к 1, сводеш можно определить как расстояние между двумя изолектами при несовпадении их основных списков на 1%, а для 100-словного основного списка – на 1 слово. Заменяя значение $c < 1$ на более удобное (целое) значение $C = 100c$, расстояние между двумя изолектами (в сводешах) определяется по формуле

$$L = -100 \ln \frac{C}{100} = 100 \ln \frac{100}{C}. \qquad (1)$$

Рассмотрим наиболее типичные случаи расстановки определенного количества современных языков и методику определения соответствующих ретроспективных связей. В общем случае для $k$ языков имеем $n = \frac{k(k-1)}{2}$ независимых значений коэффициента $C$.

**Случай 1.** Два языка. Одно независимое значение коэффициента совпадений. Соответствующая случаю матрица коэффициентов совпадений приведена на рисунке 1.

| Язык | 1 | 2 |
|------|---|---|
| 1 | – | $C_{12}$ |
| 2 | $C_{12}$ | – |

Рис. 1

| Язык | 1 | 2 | 3 |
|------|---|---|---|
| 1 | – | $C_{12}$ | $C_{13}$ |
| 2 | $C_{12}$ | – | $C_{23}$ |
| 3 | $C_{13}$ | $C_{23}$ | – |

Рис. 2

Расстояние между языками $L_{12} = 100\ln\dfrac{100}{C_{12}}$. Извлечь какую-либо дополнительную информацию из этого значения невозможно. Рассматриваемые языки могли явиться результатом дивергенции общего изолекта-предка, существовавшего $\dfrac{L_{12}}{2}$ сводешей тому назад, либо находятся в соотношении «лексификатор – пиджин», т.е. один из языков можно рассматривать как результат современного нам скрещивания другого языка с гипотетическим (не рассматриваемым нами) третьим языком. Возможны промежуточные варианты: оба языка дивергировали $L'_{12} < L_{12}$ сводешей тому назад из крайних точек изолектной цепи шириной $(L_{12} - 2L'_{12})$ сводешей.

**Случай 2.** Три языка. Три независимых значений коэффициента совпадений. Соответствующая случаю матрица коэффициентов совпадений приведена на рисунке 2.

Из трех значений коэффициентов совпадений ($C_{12}$, $C_{13}$ и $C_{23}$) выберем наибольший, соответствующий наиболее близким языкам. Не теряя общности рассуждений, будем считать наибольшим значением $C_{12}$. Соотношение между языками 1 и 2 в наиболее общем виде отображено на рис. 3. В данной работе на дендрограммах языки отображаются ромбами, узлы дендрограмм (изолекты) – точками. Языки и изолекты отмечены числами, выделенными жирным шрифтом. Числа у звеньев означают их длину в сводешах.

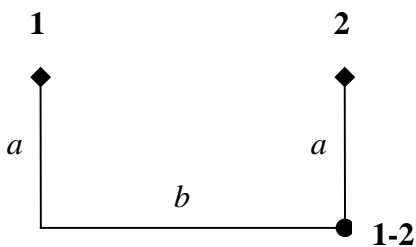

Рис. 3

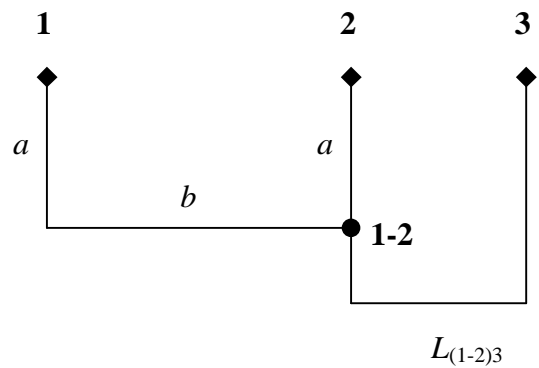

Рис. 4

Языки дивергируют из разных точек существовавшей $a$ сводешей тому назад изолектной цепи шириной $b$ сводешей. Существует условие $L_{12} = 2a + b$, откуда $a = \frac{L_{12} - b}{2}$. В общем случае один из двух языков (1 или 2) располагается ближе к третьему языку (3). В зависимости от знака разности $(L_{13} - L_{23})$ язык 3 подключен к системе 1-2 через левую либо правую точку изолектной цепи шириной $b$. Опять-таки, не теряя общности, полагаем, что этой точкой является узел 1-2. Разность расстояний между языками 1 и 3 и 2 и 3 составляет согласно рисунку 3 $\Delta L = L_{13} - L_{23} = b$, что позволяет определить $b = L_{13} - L_{23}$ и $a = \frac{L_{12} - L_{13} + L_{23}}{2}$. На следующем этапе определяем расстояние от языка 3 до узла 1-2. Это расстояние можно определить двумя способами: с отсчетом от языка 1 ($L^{(1)}_{(1-2)3}$) или от языка 2 ($L^{(2)}_{(1-2)3}$):

$$L^{(1)}_{(1-2)3} = L_{13} - (a+b) = L_{13} - \frac{L_{12} - L_{13} + L_{23}}{2} - (L_{13} - L_{23}) = \frac{L_{23} - L_{12} + L_{13}}{2};$$

$$L^{(2)}_{(1-2)3} = L_{23} - a = L_{23} - \frac{L_{12} - L_{13} + L_{23}}{2} = \frac{L_{23} - L_{12} + L_{13}}{2}.$$

Обратим внимание, что для случая трех языков оба значения совпадают. В отношении последнего подключаемого звена существует неопределенность, связанная с тем, что конфигурация звена должна определяться по еще не подключенным звеньям, каковые в случае последнего подключаемого звена отсутствуют. В то же время известное значение $L_{(1-2)3}$ ограничивает предельно возможную глубину звена, которая не может быть больше чем $\frac{L_{23}}{2}$. Дендрограмма системы 1-2-3 приведена на рисунке 4.

**Случай 3.** Четыре языка. Шесть независимых значений коэффициента совпадений. Общий расчет системы подобной сложности, а также более сложных затруднен ввиду резкого увеличения вариантов взаимного расположения языков и изолектов-предков, однако расчет системы любой сложности с конкретными значениями коэффициентов совпадений не представляет трудности. На каждом этапе конструирования дендрограммы выбирается пара точек (языков или узлов) с минимальным взаимным расстоянием. Расчет ширины изолектных цепей производится с учетом разницы расстояний от двух рассматриваемых точек до всех выявленных узлов и еще не включенных в дендрограмму языков. Поскольку при сложных системах возможно противоречие между отдельными данными вследствии промахов при измерениях, неполноты данных, неполного соответствия модели языковой действительности и пр., эти противоречия считаются носящими случайный характер (т.е. флюктуациями) и их действие сводится к минимуму путем вычисления математических ожиданий параметров системы. Возникающие «невязки» согласно общепринятым правилам измерений «раскидываются» по элементам конструкции (звеньям дендрограммы). После конструирования дендрограммы необходимо просчитать попарно расстояния между всеми языками (путем суммирования длин отдельных звеньев) и по формуле

$$C = 100\exp\left(-\frac{L}{100}\right) \quad (2)$$

найти соответствующие теоретическим (согласно модели дендрограммы) расстояниям между языками теоретические коэффициенты совпадения. Сравнение их с измеренными позволяет судить о степени адекватности модели языковой действительности.

Наша работа позволяет найти подходы к поставленному В.А. Звегинцевым вопросу о датировке распада языковых групп, представленных значительным количеством языков (например, славянских и балтийских, славянских и германских и т.д.) [3, с. 19]. Предлагаемый метод позволяет построить генеалогическое дерево языков рассматриваемой группы, но принципиально неспособен указать точку связи рассматриваемой системы с внешней системой (для чего нужен учет внешних связей). В то же время метод выявляет "кандидатов" на прямое родство с общим (единственным) языком-предком. Это языки, лежащие на линиях дивергентного развития, исходящих из точек диалектной цепи языка-предка данной группы, и не подвергавшиеся скрещиванию за период, отражаемый дендрограммой данной группы.

Под отсутствием скрещивания в рамках модели понимается отсутствие такового в соответствии с дендрограммой, построенной для языковой группы данного состава. (Не исключено, что при расширении списка обследованных языков языки, ранее рассматривавшиеся как не подвергавшиеся скрещиванию, потеряют свой статус). При установлении общего родства двух или нескольких языковых групп необходимо ограничится отбором от каждой группы именно этих выявленных "кандидатов". Для балтославянской языковой подсемьи таковыми являются прусский, литовский, белорусский и словацкий языки [2]. Один из этих языков и является современным потомком гипотетического балтославянского праязыка предыдущего уровня (или ближе всех рассматриваемых языков к нему, если прямой потомок отсутствует или не привлечен к рассмотрению).

В некоторых работах предложено устранять заимствования из основных списков, поскольку считается, что замены в пределах исконной лексики являются равномерным процессом, а заимствования – случайными сбоями, носящими нерегулярный характер [4, с. 10, 32]. В результате исключения из основных списков заимствований (вопроса надежного разделения заимствований и замен мы здесь не касаемся) обработке подверга-ются завышенные значения коэффициентов совпадений, что вносит систематические искажения в получаемые результаты. Проиллюстрируем сказанное примером. Имеем языковую систему согласно матрице рисунка 2, которой соответствует дендрограмма рисунка 4. Будем считать, что основной список каждого из трех языков системы содержит $N_з$ заимствований, причем во всех языках на заимствования заменялись одни и те же элементы списка. При исключении заимствований из основных списков размером $N_0$ основные списки сократятся до ($N_0 - N_з$) слов, а коэффициент совпадения между языками $m$ и $n$ изменится и составит $C'_{mn} = C_{mn} \frac{N_0}{N_0 - N_з}$. Вычисленное по формуле (1) расстояние также изменится:

$$L'_{mn} = 100 \ln \frac{100}{C'_{mn}} = 100 \ln \left( \frac{100}{C_{mn}} \cdot \frac{N_0 - N_з}{N_0} \right) = 100 \ln \frac{100}{C_{mn}} + 100 \ln \frac{N_0 - N_з}{N_0} =$$

$$= L_{mn} - 100 \ln \frac{N_0}{N_0 - N_з}.$$

Итак, в рассматриваемом примере все расстояния между языками уменьшаются на одну и ту же величину $s = 100 \ln \frac{N_0}{N_0 - N_з}$. Для 100-словного основного списка при малости $N_з$ выполняется приближенное соотношение $s \approx N_з$. Значение ширины изолектной цепи $b = L_{13} - L_{23}$ в результате исключения заимствований не изменится:

$b' = L'_{13} - L'_{23} = L_{13} - s - L_{23} + s = L_{13} - L_{23} = b$. Вертикальные составляющие (линии дивергенции) при пересчете изменятся:

$$a' = \frac{L'_{12} - L'_{13} + L'_{23}}{2} = \frac{L_{12} - s - L_{13} + s + L_{23} - s}{2} = \frac{L_{12} - L_{13} + L_{23} - s}{2} = a - \frac{s}{2}.$$

Поскольку $L_{(1-2)3}$ также уменьшится на величину $\frac{s}{2}$, налицо уменьшение всех вертикальных составляющих дендрограммы. Расчет системы в наиболее общем случае (при условии, что заимствования присутствуют в списке лишь одного языка, условии неравенства количества заимствований либо замены на заимствования несовпадающих элементов списка) показывает, что вертикальные составляющие изменяются всегда в сторону уменьшения, ширина же изолектных цепей может как уменьшаться, так и увеличиваться.Уменьшение вертикальных составляющих (т.е. линий дивергентного развития, отражающих время самостоятельного развития языков) и деформация ширины изолектных цепей искажает дендрограмму и побуждает вводить вместо действительного значения скорости изменения лексики измененное (уменьшенное) значение (таблица в [4, с. 10]).

Поясним сказанное. Из предложенной Сводешом формулы $t = -\dfrac{\ln c}{\lambda}$ и выражения $L = -100\ln c$ вытекает $t = \dfrac{L}{100\lambda}$, где $t$ – время развития в тысячелетиях. Между длиной вертикальных составляющих, измеренной в сводешах, и длительностью процесса, отражаемого длиной линии дивергентного развития, существует прямая зависимость (пропорциональность) – рисунок 5, кривая 1. Если следовать методике исключения заимствований из подсчета, зависимость 1 сдвигается влево на $s$ (зависимость 2).

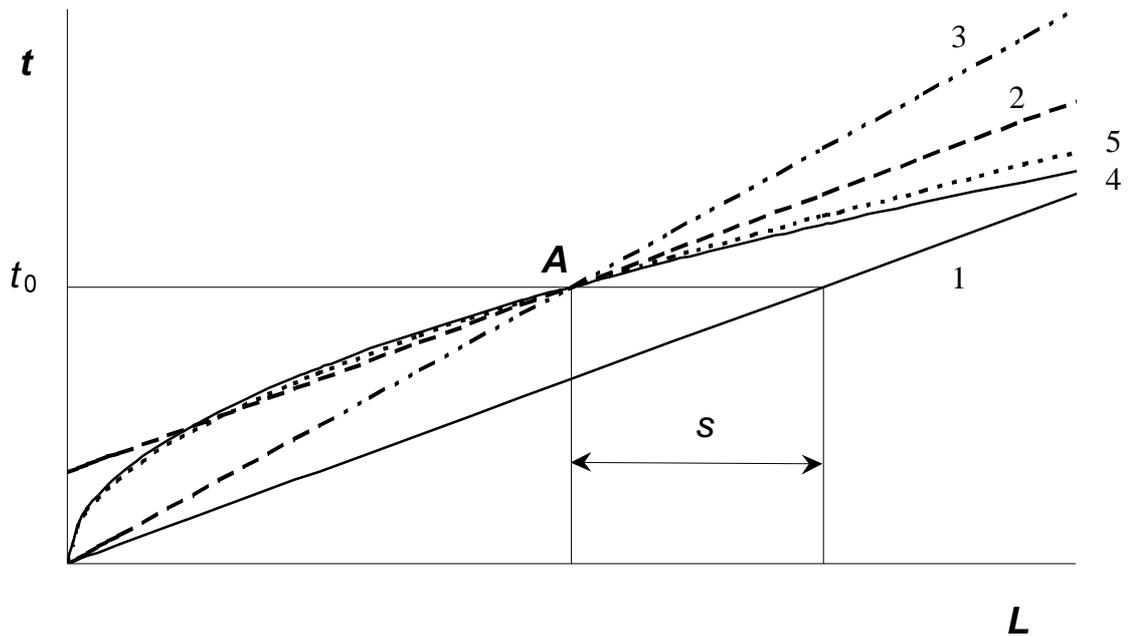

Рис. 5

Через какую-либо точку прямой 2 (например, $A$, соответствующую определенному времени развития $t_0$, например 1 тысячелетию) прямая 3 из начала координат может быть проведена лишь при условии уменьшения постулируемой скорости изменения лексики $\lambda_2 < \lambda_1$. Из рисунка 5 вытекает "умоложенная" датировка недавних лингвистических событий и удревнение датировок отодвинутых по времени событий (примеры приведены в [4, с. 11]).

Переход к формуле (5) "равномерно ускоренного движения" $c = e^{-\lambda t^2}$ исходя из предположения "о старении слов" [4, с. 13] (или $t = \sqrt{\dfrac{L}{100\lambda}}$) соответствует кривой 4 на рисунке 5. Из рисунка следует, что эта зависимость лучше, чем зависимость 3, отражает действительность в области значений $t < t_0$ (последние тысяча – полутора тысяч лет [4, с. 14]), но на больших временных глубинах начинает давать значительное умоложение датировок [4, с. 14] по сравнению с зависимостью 2. Ревизия формулы (5) из [4, с. 13] с поправкой на переменную скорость распада отдельных частей основного списка, зависящую от доли сохранившихся слов, ведет к формуле (9) из [4, с. 16]: $c = e^{-\lambda c t^2}$, откуда следует $t = e^{0,005L}\sqrt{\dfrac{L}{100\lambda}}$ (кривая 5 на рисунке 5). Дополнительный член $e^{0,005L}$ удревняет датировки при увеличении $t$, что еще больше сближает кривые 5 и 2 (рисунок 5).

Не отрицая в целом эффект "старения" слов и разную скорость распада отдельных частей основного списка, мы считаем, что в формуле $c = e^{-\lambda c t^2}$ влияние этих эффектов гипертрофировано по сравнению с языковой действительностью, что вызывается необходимостью компенсировать эффекты, вызванные неучетом заимствований. Продифференцируем более общую, чем (9) из [4, с. 16] формулу: $c = e^{-\lambda c t^\alpha}$. $\dfrac{dc}{dt} = -\lambda c \alpha t^{\alpha-1} e^{-\lambda c t^\alpha}$. Лишь значение $\alpha = 1$ обеспечивает приемлемое значение производной (т.е. скорости изменения основного списка) при $t = 0$, поскольку при $\alpha > 1$ скорость равна 0, а при $\alpha < 1$ – бесконечности. При наличии феномена увеличения скорости распада в более ранние периоды, зависимость $c(t)$ должна исключать нереальные значения производной в точке $t = 0$. Подобную зависимость еще предстоит найти.

Вышеприведенные выкладки и рисунок 5 приведены с целью иллюстрации того факта, что одна и та же языковая действительность может быть удовлетворительно описана исходя из противоположных теоретических воззрений путем конструирования подходящих формул. Наиболее приемлемой с практической точки зрения оказывается зависимость с большей областью применения и большей прогностической способностью.

В качестве примера применения описанной методики произведем построение дендрограммы индоарийских языков. Используем приведенные в работе [5] данные по диалектам европейских цыган (кэлдэрарскому, северно-русскому и финскому) и языкам Индостана (хинди, панджаби и непали). Из подсчета исключаем язык армянских цыган (боша) и язык гиссарских парья, поскольку данных по этим языкам недостаточно для параметризации в рамках модели. Для привлеченных к обследованию шести языков по данным Списка № 1 и Списка № 2 [5, с. 34–39] составлен единый основной список объемом $N_0 = 94$ слова. В таблице 1 приведены коэффициенты совпадений, использованные при расчетах.

Таблица 1

| № | Язык | 1 | 2 | 3 | 4 | 5 | 6 |
|---|---|---|---|---|---|---|---|
| 1 | Кэлдэрарский | – | 79 | 80 | 53 | 52 | 50 |
| 2 | Северно-русский | 79 | – | 80 | 52 | 52 | 50 |
| 3 | Финский | 80 | 80 | – | 54 | 53 | 51 |
| 4 | Хинди | 53 | 52 | 54 | – | 79 | 63 |
| 5 | Панджаби | 52 | 52 | 53 | 79 | – | 65 |
| 6 | Непали | 50 | 50 | 51 | 63 | 65 | – |

Построенная дендрограмма приведена на рисунке 7. Опять-таки, конфигурация последнего подключаемого звена между узлами 1-3 и 4-6 не может быть определена без привлечения данных о внешних связях языковой системы. Известна лишь длина звена – 32 сводеша. На дендрограмме изображены два возможных предельных варианта конфигурации звена – в виде диалектной цепи шириной 23 сводеша (первый вариант) и в виде звена с глубоко лежащей исходной точкой дальнейшего развития О (второй вариант).

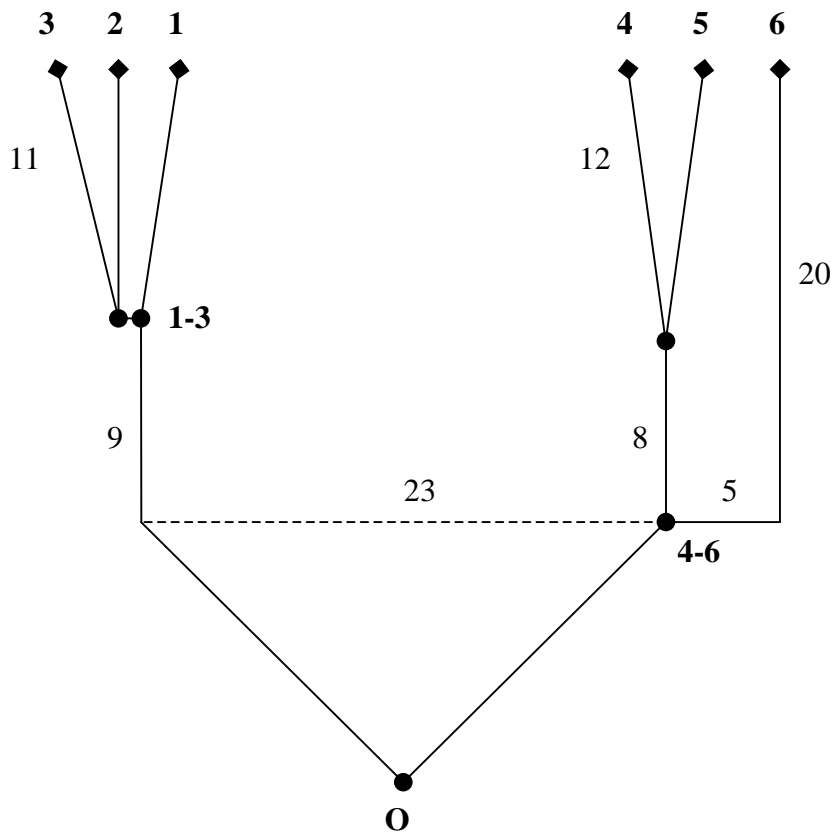

Рис. 7

Дендрограмма рисунка 7 построена с соблюдением масштаба по вертикали и горизонтали. Из дендрограммы следует, что единый предок рас-сматриваемых языков существовал около 31–32 сводешей тому назад (рассматривается второй вариант). Распад этого языка-предка дал начало ветвям языков Индостана и языка-предка цыганских диалектов. Через 11-12 сводешей произошло отделение языка непали (на базе диалектной цепи шириной 5 сводешей), а еще через 8–9 сводешей начали самостоятельно развиваться хинди и панджаби и цыганские диалекты.

При исключении из подсчетов заимствований получаем таблицу 2 с новыми коэффициентами совпадений и дендрограмму рисунка 8. Ввиду малого размера изображения частной дендрограммы цыганских диалектов эта часть общей дендрограммы изображена в правом нижнем углу рисунка 8 в увеличенном масштабе.

Таблица 2

| № | Язык | 1 | 2 | 3 | 4 | 5 | 6 |
|---|---|---|---|---|---|---|---|
| 1 | Кэлдэрарский | – | 95 | 95 | 77 | 67 | 62 |
| 2 | Северно-русский | 95 | – | 96 | 75 | 67 | 60 |
| 3 | Финский | 95 | 96 | – | 78 | 68 | 63 |
| 4 | Хинди | 77 | 75 | 78 | – | 95 | 76 |
| 5 | Панджаби | 67 | 67 | 68 | 95 | – | 73 |
| 6 | Непали | 62 | 60 | 63 | 76 | 73 | – |

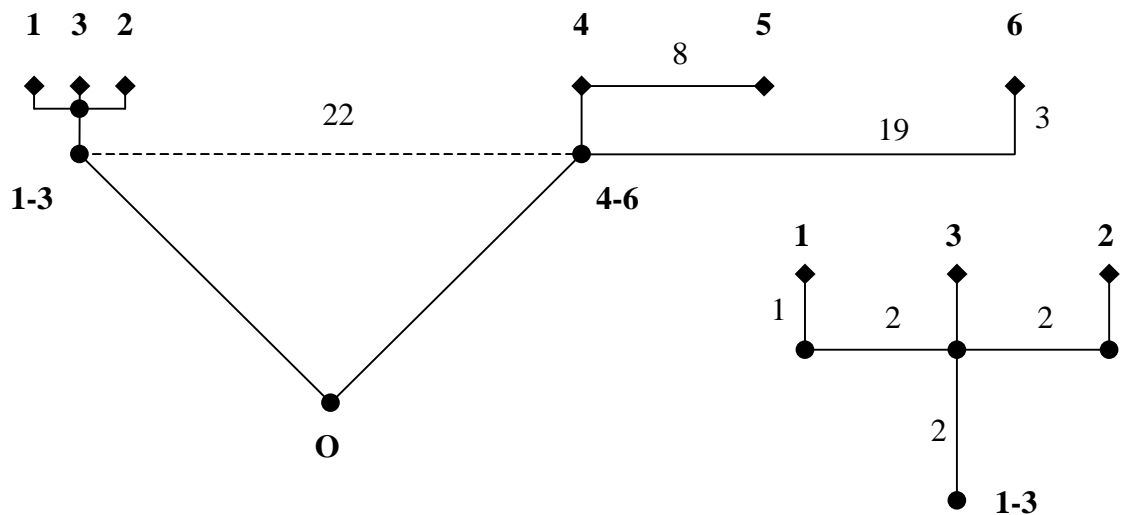

Рис. 8.

Сравнение дендрограмм рисунков 7 и 8 подтверждает высказанное и доказанное аналитически предположение, что исключение из подсчета заимствований ведет к укорочению дендрограммы по вертикали. Кроме того, существенно увеличивается ширина диалектной цепи языка непали с 5 до 19 сводешей; языки хинди и панджаби, согласно дендрограмме рисунка 7 разделившиеся 12 сводешей тому назад, на дендрограмме рисунка 8 представляются современной диалектной цепью шириной 8 сводешей.

**Библиографический список**

1. Сводеш М. Лексикостатистическое датирование доисторических этнических контактов (на материале племен эскимосов и североамериканских индейцев) // Новое в лингвистике. Вып. 1. М.: Изд-во иностр. литер., 1960. С. 23–52.
2. Kromer V. Glottochronology and problems of protolanguage reconstruction. http://www.arxiv.org/pdf/cs.CL/0303007.
3. Звегинцев В.А. Лингвистическое датирование методом глоттохронологии (лексикостатистики) // Новое в лингвистике. Вып. 1. М.: Изд-во иностр. литер., 1960. С. 9–22.
4. Старостин С.А. Сравнительно-историческое языкознание и лексикостатистика // Лингвистическая реконструкция и древнейшая история Востока (Материалы к дискуссиям международной конференции). Т. 1. М.: Наука, 1989. С. 3–39.
5. Дьячок М.Т. Глоттохронология: пятьдесят лет спустя // Сибирский лингвистический семинар. 2002. № 1. Новосибирск: Новосибирское книжное изд-во.